# Shiva++: An Enhanced Graph based Ontology Matcher

Iti Mathur, Nisheeth Joshi
Apaji Institute
Banasthali University
Rajasthan, India

Hemant Darbari, Ajai Kumar
Applied Artificial Intelligence Group
Center for Development of Advanced Computing
Pune, Maharashtra, India

## ABSTRACT

With the web getting bigger and assimilating knowledge about different concepts and domains, it is becoming very difficult for simple database driven applications to capture the data for a domain. Thus developers have come out with ontology based systems which can store large amount of information and can apply reasoning and produce timely information. Thus facilitating effective knowledge management. Though this approach has made our lives easier, but at the same time has given rise to another problem. Two different ontologies assimilating same knowledge tend to use different terms for the same concepts. This creates confusion among knowledge engineers and workers, as they do not know which is a better term then the other. Thus we need to merge ontologies working on same domain so that the engineers can develop a better application over it. This paper shows the development of one such matcher which merges the concepts available in two ontologies at two levels; 1) at string level and 2) at semantic level; thus producing better merged ontologies. We have used a graph matching technique which works at the core of the system. We have also evaluated the system and have tested its performance with its predecessor which works only on string matching. Thus current approach produces better results.

## 1. INTRODUCTION
In current information age, we are experiencing information/knowledge explosion. Often we come across information or knowledge sources (Ontologies) which have similar concepts but have no interoperability between them. Some sources capture a part of the information in detail while others may capture some other information. Thus, in order to have a complete understanding of the concept the user has to browse through all these sources. At times this creates confusion as different sources tend to use different terminology for different concepts.

Thus, we require a mechanism which can merge these different representations of similar concepts into one. This is achieved by ontology matching. At the core the basic objective of ontology matching is to provide heterogeneous information into one common platform. Ideally, this common platform produces results into five tuples <*id, e, e′, r,n*> for two ontologies *o* and *o′*. Here *id* is the identifier which represents the correspondence of concepts between two ontologies. *e* and *e′* are elements of *o* and *o′* respectively. *r* is the alignment relation between two elements (Similar, General, Isa etc.) and *n* is a numeric value for the alignment relation. Ideally this has some value between 0 and 1 (both inclusive).

The rest of the paper is organized as follows: Section 2 gives a brief overview of some of the ontology matchers. Section 3 gives a description of our approach. Section 4 describes the evaluation process incorporated and Section 5 concludes the paper.

## 2. LITERATURE SURVEY
Ontology matching, unlike other areas of computer science is still an unexplored territory. Though a lot of matchers have been proposed in the literature, only a handful of them have been pursued for further enhancements of both; the matching problem at large and the performance of the specific systems. In this section we shall provide a brief description of systems which have seen enhancements over time and would not be considering the ones which were used once upon a time but its current development is dormant.

Cruz at el. [1] have developed a matchers named Agreement Maker. This is considered as the best matchers as it has a very good user interface and a flexible architecture. This matchers involves the users of the system into matching process. Thus produces better results than any other matcher. The developers of this system believe that "users can help make better alignments which are not possible in automatic alignments." Thus they prophesize the use of having semi-automatic matching systems. Ruiz and Grau [2] have developed LogMap at University of Oxford. This matcher incorporates logic based reasoning approach in their matcher. Since long ontologies have used description logic to reason out new concepts. Using it in matching process is a very intuitive approach as it can produce better alignments. Though commenting on it is very early as the matcher is still in development stage and is yet to produce good results.

Jérôme [3] has developed a hybrid ontology matcher which can match the concepts and properties from two ontologies. He has used association rule paradigm [4] and statistical interestingness measure for implementing this matcher. Jorge et al. [5] have developed a matcher which tries to align ontologies using schema matching. For this, they applied two approaches, at first they extracted similar concepts and then applied different matching techniques onto the concepts extracted and finally produced aligned ontology. Peng et al. [6] have developed Lily, which is a very good matching system. It matches general and heavy-weight ontologies and produce good results for decent size ontologies, but it takes a lot of time to do so. At the core this matcher extracts semantic sub graphs and then tries to align it with other ontologies.

Juanzi [7] has developed RiMOM Ontology matcher which is one the top performing matchers being tested in various evaluation campaigns, across the globe. It is considered as a good matcher as it matches schema and instances available in the ontologies and uses multiple techniques to implement. Moreover to improve the results it also uses several external resources like WorldNet to do semantic matching. Fayçal et al. [8] have developed TaxoMap ontology matching system. This matcher can merge heavy-weight ontologies. It does so by finding correspondence between the concepts of two ontologies by applying subsumption, inverse and proximity relations. YAM++ [9] is another system which incorporates different matching algorithms which are combined to produce merged ontology. This system is self-configurable and extensible, as if the user is not satisfied with the results then he can provide his own customized matching approach. Mathur et al. [10] have developed a graph based ontology matcher which can use any one of the string matching algorithm to be combined with bi-partite graph matching.

## 3. OUR APPROACH

### 3.1 Experimental Setup

In order to experiment with our matcher, we have used some of the ontologies with OAEI (Ontology Alignment and Evaluation Initiative) 2013 evaluation task [11]. This task had some lightweight ontologies and one heavyweight ontology. We used six ontologies from benchmark test set. These were light weight ontologies. We also used an ontology from anatomy track.

Since this was inadequate, as this evaluation campaign only uses one heavy-weight ontology and needed more such ontologies to make a sound assumption, so, we developed some ontologies on our own. We have developed three ontologies; a ontology on human anatomy [12] which had concepts relating to human physiological structure; a ontology on health care services [13] and a ontology on communicable diseases [14].

In order to validate our results we used RiMOM and YAM++ ontology matchers and compared our system with them. We used a graph based approach with Word Net for matching the ontologies. The objective was to enhance the capability of our graph matching ontology matcher, as it has produced some promising results.

### 3.2 Methodology

This system is an enhancement over our previous graph based ontology matching system, "Shiva". Thus we have named it Shiva++. As its processor it uses graph matching algorithm at its core which is wrapped up by string and semantic matching algorithms and resources.

In our approach, we first take two ontologies. Since we support heterogeneity, the two ontologies need not be in same format. Our system is capable of recognizing different formats and extract concepts, sub-concepts, and attributes from ontologies. For our purpose, the two ontologies are termed as the source ontology $O_s$ and target ontology $O_t$. These are first read and are sent for preprocessing where both the ontologies are separately parsed and concepts, sub-concepts, properties and instances of both the ontologies are collected separately. This extracted information is then arranged in a linked graph in memory.

After preprocessing, the extracted information is sent to the matching system, here the user can choose between four String matching algorithms, these are: Levensthein Edit Distance [15], Qgrams [16], Smith Waterman [17] and Jaccard's Coefficient [18] algorithms. The selected algorithm matches the similarities between concepts, sub-concepts, properties and instances. If some of the elements (concepts, sub-concepts, properties and instances) are not matched by the algorithm then they are passed to the word net for semantic similarity. This matching can not only match the synonymous elements but can also match general with specific elements. For example, in an ontology a concept is named as "car" which has information about different cars and in another ontology a concept is named as "vehicle" which has similar information. A simple string matcher cannot match these two concepts. Even just synonymous matches would not work. But by using word net, we can get the complete hierarchical structure of the car and there, car can be matched with vehicle.

Using these matches we generated a score matrix in the following format:

$$S = \begin{bmatrix} M[O_{11}O_{21}] & M[O_{12}O_{21}] & M[O_{13}O_{21}] \dots M[O_{1m}O_{21}] \\ M[O_{11}O_{22}] & M[O_{12}O_{22}] & M[O_{13}O_{22}] \dots M[O_{1m}O_{22}] \\ M[O_{11}O_{23}] & M[O_{12}O_{23}] & M[O_{13}O_{23}] \dots M[O_{1m}O_{23}] \\ \vdots & \vdots & \vdots & \vdots \\ \vdots & \vdots & \vdots & \vdots \\ M[O_{11}O_{2n}] & M[O_{12}O_{2n}] & M[O_{13}O_{2n}] \dots M[O_{1m}O_{2n}] \end{bmatrix}_{m \times n}$$

Here, $M[O_{11}O_{21}]$ is the mapping between one of the elements (concepts, sub-concepts, properties, instances) of source ontology $O_s$ with one of the elements (concepts, sub-concepts, properties, instances) of target ontology $O_t$. Each element has the value produced by either of the similarity metrics. For example, if we are using levensthein distance algorithm and we have two concepts as "pizza" and "pizzas", then its score would be 1 and the similarity is calculated using the formula in equation 1.

$$sim(x, y) = \frac{\#edits\,(x,y)}{\max\{len\,(x), len\,(y)\}} \qquad (1)$$

Here x and y are the two strings, in our case x is "pizza" and y is "pizzas". #matches(x,y) is the no. of edits required to make the two strings equal and len(x) is the length of string x, len(y) is the length of string y. The maximum of the two is selected to compute the final score.

Similarly if the two concepts are "pizza" and "food". Then no string matching algorithm can match these two. But if we use external ready-made resources which can provide the general terms for a term, then we can match the two as food is a more general representation of pizza. Thus the similarity of these two terms is given a score "0.7".

This process is repeatedly done for all the matches between the two ontologies. Thus this finally generates the score matrix of all the matched elements. This matrix can be seen as bipartite graph which has two disjoint sets of vertices (in our case mapping elements of $O_s$ and $O_t$) with edge weights (similarity values) clearly mentioned.

This score matrix is then used with the graph matching algorithm. We have used Hungarian method [19] for matching our score matrix (bipartite graph). The entire working of the system is depicted in Figure 1. Figure 2 shows a snapshot of the merged ontology.

## 4. EVALUATION

To evaluate the performance of our system we used 10 ontologies. Among them 6 were light weight ontologies and 4 were heavy weight ontologies. We have used 2 popular ontology matchers (RiMOM and YAM++) with our two matchers (Shiva and Shiva++). For this study we have restricted our experiments to levensthein's edit distance algorithm with word net (only for Shiva++). We calculated precision, recall and f-measures using equations 2,3 and 4 respectively.

$$Precision\,(P) = \frac{\#correct\_mappings}{\#total\_mappings\_system} \qquad (2)$$

$$\text{Recall (R)} = \frac{\#correct\ \_mappings}{\#total\_mappings\ \_human} \qquad (3)$$

$$F - Measure\ (F) = \frac{2 \times P \times R}{P + R} \qquad (4)$$

Table 1 shows the values of precision, recall and fmeasure. Among these ontology nos. 1-6 were light-weight ontologies and ontology nos. 7-10 were heavy-weight ontologies. For ontology nos. 1-3, RiMOM had the best f-measure socre, while for ontology nos. 4-5, YAM++ had the best score. In all these cases Shiva++ was trailing behind the leaders. For ontology nos. 6-10, Shiva++ had the best score.

## 5. CONCLUSION

In this paper, we have shown the implementation of an enhanced graph based matcher which works on string similarity and semantic similarity. This was coupled with bipartite graph matching algorithm which created matched ontology. This approach produced good results; with heavyweight ontologies it produced the best results while with lightweight ontologies it could produce moderate results. Out of the six lightweight ontologies, it was able to produce best scores for just one matching. It was trailing behind either RiMOM or YAM++ which were the best matchers for these ontologies. For heavy-weight ontologies, Shiva++ was the best matcher for all four heavy-weight ontologies.

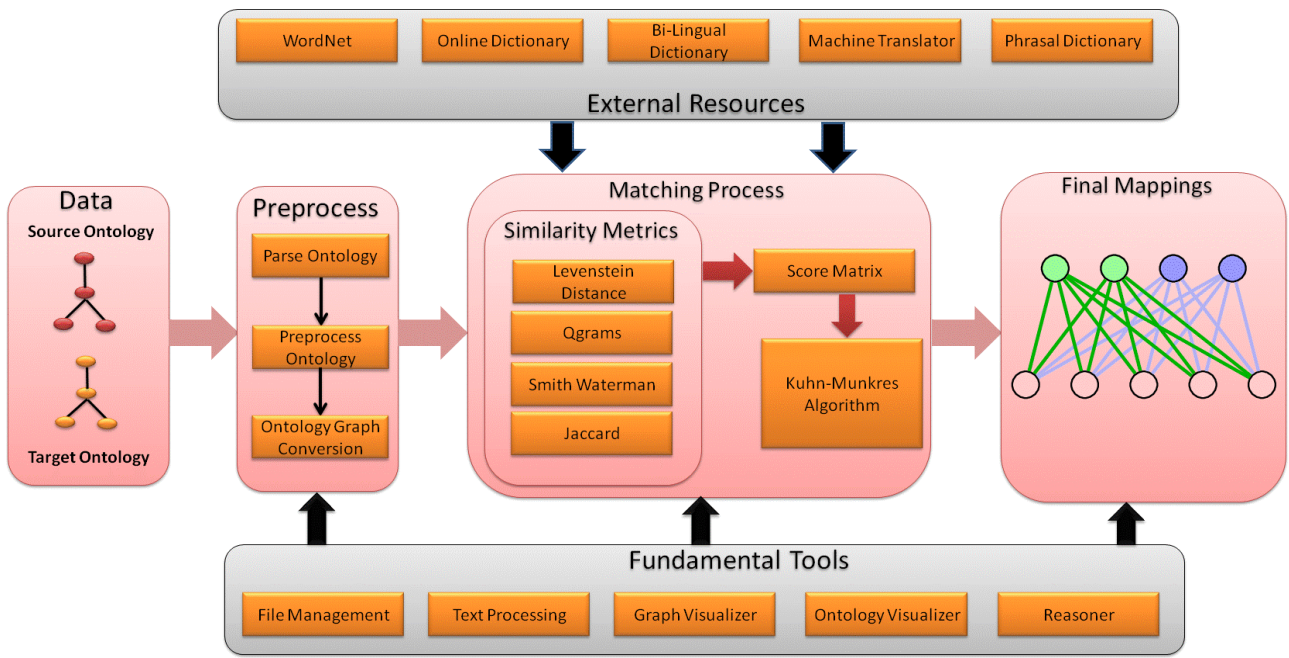

**Figure 1: Architecture of Shiva++ Ontology Matching System**

```
− <rdf:RDF>
  − <Alignment>
      <xml>yes</xml>
      <level>0</level>
      <type>11</type>
  + <onto1></onto1>
  + <onto2></onto2>
  + <uri1></uri1>
  + <uri2></uri2>
  − <map>
    − <Cell>
        <entity1 rdf:resource="http://oaei.ontologymatching.org/2011/benchmarks/101/onto.rdf#type"/>
        <entity2 rdf:resource="http://oaei.ontologymatching.org/2011/benchmarks/101/onto.rdf#type"/>
        <measure rdf:datatype="http://www.w3.org/2001/XMLSchema#float">1.0</measure>
        <relation>=</relation>
      </Cell>
    </map>
  − <map>
    − <Cell>
        <entity1 rdf:resource="http://oaei.ontologymatching.org/2011/benchmarks/101/onto.rdf#Periodical"/>
        <entity2 rdf:resource="http://oaei.ontologymatching.org/2011/benchmarks/101/onto.rdf#Journal"/>
        <measure rdf:datatype="http://www.w3.org/2001/XMLSchema#float">0.7</measure>
        <relation>=</relation>
      </Cell>
    </map>
```

**Figure 2: Snapshot of aligned ontology**

| Ontology No. | | RiMOM | | | YAM++ | | | Shiva | | | Shiva++ | | |
|---|---|---|---|---|---|---|---|---|---|---|---|---|---|
| | | **P** | **R** | **F** | **P** | **R** | **F** | **P** | **R** | **F** | **P** | **R** | **F** |
| **Light-weight Ontologies** | 1 | 1 | 0.989691 | **0.994819** | 0.968085 | 0.938144 | 0.95288 | 1 | 0.987952 | 0.993939 | 1 | 0.98913 | 0.994536 |
| | 2 | 0.96875 | 0.958763 | **0.963731** | 0.723404 | 0.701031 | 0.712042 | 0.967033 | 0.956522 | 0.961749 | 0.963415 | 0.963415 | 0.963415 |
| | 3 | 0.871875 | 0.8628867 | **0.8673579** | 0.6510636 | 0.6309279 | 0.6408378 | 0.8670735 | 0.8670735 | 0.8670735 | 0.8670735 | 0.8670735 | 0.8670735 |
| | 4 | 0.909091 | 0.721649 | 0.804598 | 1 | 0.989691 | **0.994819** | 0.905405 | 0.72043 | 0.802395 | 0.909091 | 0.722892 | 0.805369 |
| | 5 | 0.861111 | 0.639175 | 0.733728 | 0.96875 | 0.639175 | **0.770186** | 0.855072 | 0.641304 | 0.732919 | 0.855072 | 0.641304 | 0.732919 |
| | 6 | 0.92 | 0.71134 | 0.802326 | 0.842105 | 0.494845 | 0.623377 | 0.916667 | 0.709677 | 0.8 | 0.926667 | 0.719677 | **0.8104** |
| **Heavy-weight Ontologies** | 7 | 1 | 0.824742 | 0.903955 | 0.933333 | 0.57732 | 0.713376 | 1 | 0.826087 | 0.904762 | 1 | 0.829268 | **0.906667** |
| | 8 | 1 | 0.865979 | 0.928177 | 1 | 0.917526 | 0.936989 | 1 | 0.865854 | 0.928105 | 1 | 0.917526 | **0.956989** |
| | 9 | 1 | 0.7793811 | 0.8353593 | 0.987952 | 0.845361 | 0.911111 | 1 | 0.865854 | 0.928105 | 1 | 0.869565 | **0.930233** |
| | 10 | 1 | 0.845361 | 0.916201 | 0.987179 | 0.793814 | 0.88 | 1 | 0.843373 | 0.915033 | 1 | 0.847826 | **0.917647** |

**Table 1: Evaluation Results**